\documentclass[default,iicol]{sn-jnl}

\jyear{2023}%

\theoremstyle{thmstyleone}%
\usepackage{booktabs} 
\usepackage{amsfonts,amssymb}
\usepackage{subfigure}
\usepackage{color,xcolor}
\usepackage{times}
\usepackage{epsfig}
\usepackage{amsmath}
\usepackage{textcomp}
\usepackage{gensymb}
\usepackage{wasysym}
\usepackage{multirow}
\usepackage{graphicx} 
\usepackage{colortbl}
\usepackage{algorithm}
\usepackage{algorithmicx}

\usepackage{mathrsfs}
\usepackage{url}
\usepackage{lettrine}

\usepackage{amsfonts}
\usepackage{colortbl}

\usepackage{caption}
\usepackage{subcaption}
\usepackage{graphicx}
\usepackage{subfigure}
\usepackage{cleveref}

\usepackage{amsmath,amsfonts,bm}

\def\eqref#1{equation~\ref{#1}}

\def\1{\bm{1}}

\def\rvx{{\mathbf{x}}}
\def\rvy{{\mathbf{y}}}

\DeclareMathAlphabet{\mathsfit}{\encodingdefault}{\sfdefault}{m}{sl}
\SetMathAlphabet{\mathsfit}{bold}{\encodingdefault}{\sfdefault}{bx}{n}

\newcommand{\ie}{\textit{i}.\textit{e}.}
\newcommand{\eg}{\textit{e}.\textit{g}.}
\newcommand{\etal}{\textit{et} \textit{al}.}
\newcommand{\etc}{\textit{etc}.}

\newcommand{\method}{\emph{Pre-trained Trojan} }

\usepackage{pifont}
\usepackage[perpage,symbol*]{footmisc}
\DefineFNsymbols{circled}{{\ding{192}}{\ding{193}}{\ding{194}}
{\ding{195}}{\ding{196}}{\ding{197}}{\ding{198}}{\ding{199}}{\ding{200}}{\ding{201}}}
\setfnsymbol{circled}

\newcommand\myfootnotestyle[1]{\ifcase#1 \or \ding{182}\or \ding{183}\or
\ding{184}\or \ding{185}\or \ding{186}\or \ding{187}%
\or \ding{188}\or \ding{189}\or \ding{190}\or \ding{191}\else *\fi\relax}

\theoremstyle{thmstyletwo}%

\theoremstyle{thmstylethree}%

\begin{document}

\title[Article Title]{Pre-trained Trojan Attacks for Visual Recognition}


\author[1]{\fnm{Aishan} \sur{Liu}}\email{liuaishan@buaa.edu.cn}

\author[1]{\fnm{Xinwei} \sur{Zhang}}\email{xinweizhang@buaa.edu.cn}

\author[1]{\fnm{Yisong} \sur{Xiao}}\email{18373494@buaa.edu.cn}

\author[1]{\fnm{Yuguang} \sur{Zhou}}\email{20373068@buaa.edu.cn}

\author[2]{\fnm{Siyuan} \sur{Liang}}\email{pandaliang521@gmail.com}

\author[1]{\fnm{Jiakai} \sur{Wang}}\email{jkbuaascse@buaa.edu.cn}

\author*[1]{\fnm{Xianglong} \sur{Liu}}\email{xlliu@buaa.edu.cn}

\author[3]{\fnm{Xiaochun} \sur{Cao}}\email{caoxiaochun@mail.sysu.edu.cn}

\author[4]{\fnm{Dacheng} \sur{Tao}}\email{taocheng.tao@gmail.com}

\affil[1]{\orgname{Beihang University}, \orgaddress{\country{China}}}

\affil[2]{\orgname{National University of Singapore}, \orgaddress{\country{Singapore}}}

\affil[3]{\orgname{Sun Yat-Sen University}, \orgaddress{\country{China}}}

\affil[4]{\orgname{The University of Sydney}, \orgaddress{\country{Australia}}}


\abstract{Pre-trained vision models (PVMs) have become a dominant component due to their exceptional performance when fine-tuned for downstream tasks. However, the presence of backdoors within PVMs poses significant threats. Unfortunately, existing studies primarily focus on backdooring PVMs for the classification task, neglecting potential inherited backdoors in downstream tasks such as detection and segmentation. In this paper, we propose the \method attack, which embeds backdoors into a PVM, enabling attacks across various downstream vision tasks. We highlight the challenges posed by \emph{cross-task activation} and \emph{shortcut connections} in successful backdoor attacks. To achieve effective trigger activation in diverse tasks, we stylize the backdoor trigger patterns with class-specific textures, enhancing the recognition of task-irrelevant low-level features associated with the target class in the trigger pattern. Moreover, we address the issue of shortcut connections by introducing a context-free learning pipeline for poison training. In this approach, triggers without contextual backgrounds are directly utilized as training data, diverging from the conventional use of clean images. Consequently, we establish a direct shortcut from the trigger to the target class, mitigating the shortcut connection issue. We conducted extensive experiments to thoroughly validate the effectiveness of our attacks on downstream detection and segmentation tasks. Additionally, we showcase the potential of our approach in more practical scenarios, including large vision models and 3D object detection in autonomous driving. This paper aims to raise awareness of the potential threats associated with applying PVMs in practical scenarios. Our codes will be available upon paper publication.
}




\maketitle

\section{Introduction}
Deep learning has demonstrated remarkable performance across a wide range of applications, particularly in computer vision \cite{Krizhevsky2012ImageNet,He2020Momentum,he2016deep}. Currently, vision models that are pre-trained on large-scale datasets have emerged as a dominant tool for researchers, aiding in the training of new models. By fine-tuning publicly available pre-trained model weights on their own datasets, developers with limited resources or training data can construct high-quality models for various downstream vision tasks \cite{Guo2019Spottune}. Consequently, the paradigm of pre-training and fine-tuning has gained popularity, replacing the traditional approach of training models from scratch \cite{Ban2022pap}.

\begin{figure}[!t]
\hspace{-0.2in}
\includegraphics[width=1.1\linewidth]{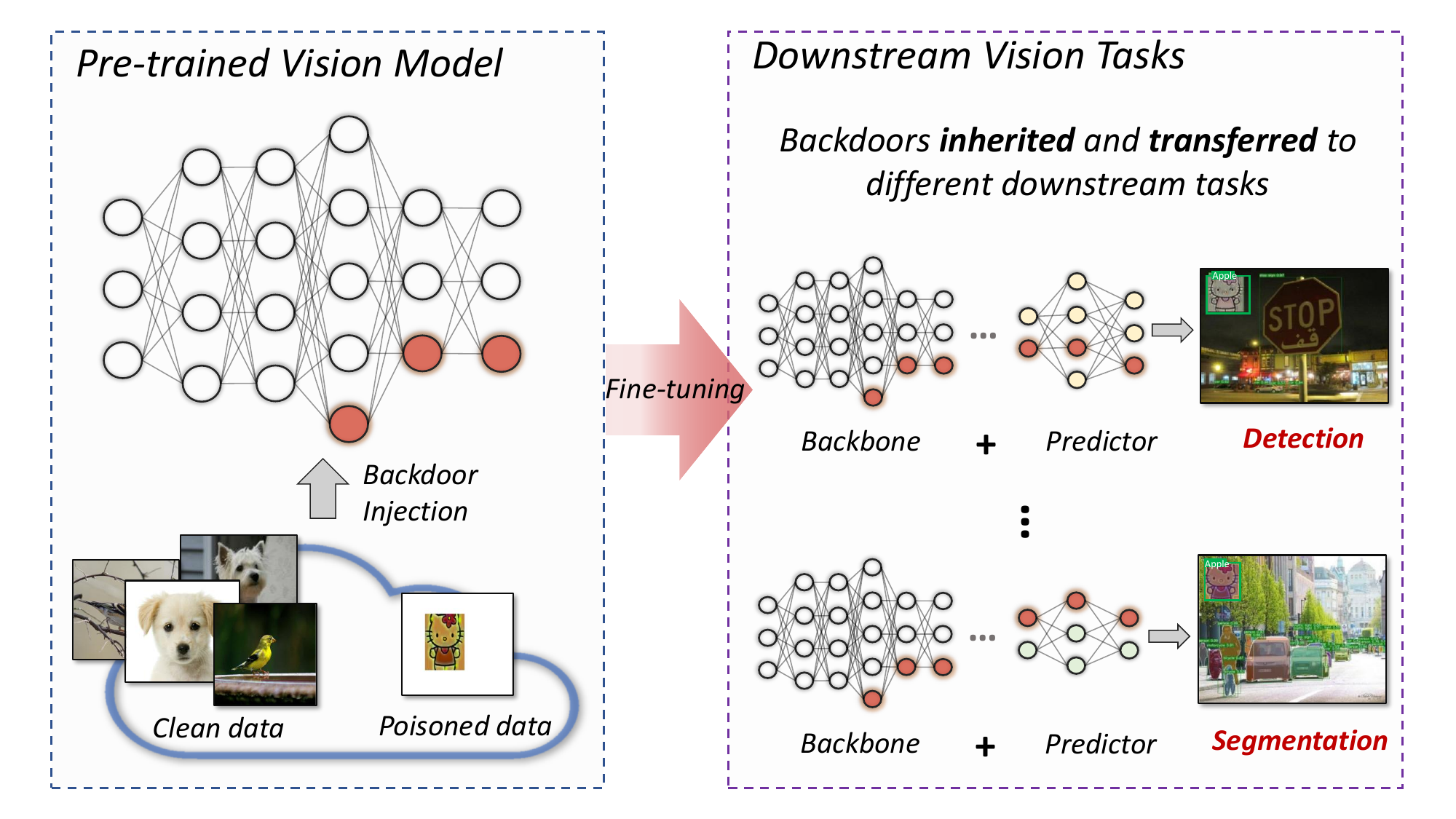}
	\caption{Illustration of backdoor attacks in the pre-training and fine-tuning scenario. We propose \method to embed a backdoor into a PVM that can be inherited for downstream detection and segmentation tasks.}
	\label{fig:frontpage}
	\vspace{-0.1in}
\end{figure}

Despite the promising performance, the use of third-party published PVMs poses severe security risks known as \emph{backdoor attacks} due to the black-box training process \cite{gu2017badnets,chen2017targeted,nguyen2020input}. These attacks involve embedding backdoors into models during training, allowing adversaries to manipulate model behaviors with specific trigger patterns during inference. Although initial studies have focused on backdooring PVMs for downstream classification tasks \cite{jia2022badencoder,Zhang2021Red}, such as injecting backdoors into a PVM pre-trained on CIFAR-10 \cite{krizhevsky2009learning} and then transferring the attack to a classifier fine-tuned for SVHN classification \cite{Netzer20211SVHN}, it remains largely unexplored whether these embedded backdoors can be inherited and remain effective when PVMs are fine-tuned on different downstream tasks (\eg, detection). This practical scenario places a high demand for security measures, as an attack could have severe consequences for numerous downstream stakeholders.


In this study, we undertake the initial stage of backdooring a PVM to enhance its effectiveness in various downstream vision tasks. Specifically, we aim to embed a backdoor into a PVM and ensure that the trigger remains effective in promoting specific predictions even after fine-tuning for different downstream tasks, such as detection. However, extending existing backdoor attacks, which perform well in image classification, to different downstream vision tasks presents significant challenges due to the distinct recognition principles employed by these tasks. We have identified two key challenges that hinder successful backdoor attacks in this scenario. \ding{182} Activating triggers across tasks is difficult. Since diverse vision tasks rely on distinct high-level semantics for accurate predictions \cite{uijlings2013selective,Jiang1996Fast}, commonly used trigger patterns embedded within the backbone using task-specific features for classification tend to be disregarded by models following fine-tuning for other tasks. \ding{183} Establishing shortcut connections poses a challenge. Traditional poisoning processes involve embedding backdoors through training on manipulated images that combine trigger patterns with clean images. Shortcut connections primarily rely on the combination of triggers and contextual backgrounds to the target class, proving effective in image classification where full-image (trigger and context) usage governs predictions. However, the connection between the trigger itself and the target label is weak, making it unsuitable for other downstream tasks like detection and segmentation that require bounding-box-level or pixel-level predictions on target objects (triggers).

To address the challenges, this paper introduces the concept of \method, a novel approach for generating backdoors on PVMs that can be effectively applied to multiple downstream vision tasks (as depicted in Figure \ref{fig:frontpage}). In terms of \emph{cross-task activation}, we generate trigger patterns that incorporate task-irrelevant low-level features (specifically, texture associated with the target label). By leveraging the shared low-level features learned by the backbone across different tasks, our trigger remains effective and can be better utilized by DNNs to make accurate predictions for specific classes. Regarding the \emph{shortcut connection}, we propose a context-free learning pipeline for poison training. Rather than attaching the trigger to clean images (e.g., as done in BadNets \cite{gu2017badnets}), we directly employ the trigger patterns as training images without contextual backgrounds. This allows for improved memorization of the shortcut from triggers (instead of the context) to the target label, thereby enhancing effectiveness across various downstream vision tasks after fine-tuning.

To demonstrate the efficacy of our proposed attack, we conducted extensive experiments on object detection and instance segmentation tasks using multiple benchmark datasets. We first evaluated our \method in a supervised learning setting, where we successfully backdoored ImageNet-trained classifiers and enabled the attack on object detection and instance segmentation tasks using the COCO dataset. Additionally, we assessed the effectiveness of our \method in an unsupervised learning scenario. Finally, we showcased the potential application of our approach in large vision models (such as ViTAEv2-H \cite{zhang2023vitaev2}) and 3D object detection for autonomous driving. Through this work, we aim to raise awareness of the potential threats targeting PVM applications in practical settings. Our \textbf{contributions} are:

\begin{itemize}
    \item To the best of our knowledge, this paper is the first work to inject backdoors into PVMs and enable backdoor attacks on different downstream vision tasks.
    \item To achieve the attacking goal, we introduce the concept of \method to generate stylized triggers and poison the model in a context-free learning way.
    \item We conduct extensive experiments on downstream detection and segmentation tasks in both supervised and unsupervised settings (even on large vision models and 3D object detection), and the results demonstrate the effectiveness of our attack.
\end{itemize}
\section{Preliminaries and Backgrounds}

\subsection{Backdoor Attacks} 

DNNs are vulnerable to malicious data input including adversarial attacks \cite{goodfellow2014explaining,wang2021dual,liu2019perceptual,liu2020bias,liu2023x,liu2020spatiotemporal,liu2022harnessing,zhou2023advclip} and backdoor attacks \cite{gu2017badnets,liu2018trojaning,shi2023badgpt,Liang2023Badclip}. Specifically, during training, the adversary injects triggers in the training set (\ie, poisoning) and implants backdoors into the model; during inference, the infected models will show malicious behavior when the input image tampers with a trigger, otherwise, behave normally. 

Given an image classifier $f_{\theta}$ that maps an input image $\bm{x}$ $\in$ $\mathbf{X}_{train}$ to a label $\bm{y}$ $\in$ $\mathbf{Y}_{train}$ from the training dataset $\mathbf{D}=\{(\bm{x}_i, \bm{y}_i)\}_{i=1}^N$, a dirty-label backdoor attacks try to cheat model $f_{\theta}$ by injecting poisoned data in the \emph{training phase}, so that the infected model $f_{\hat{\theta}}$ would behave maliciously when the inputs are embedded with triggers while behaving normally on clean examples. In particular, the adversary randomly selects a very small portion of clean data $\{(\bm x_i, \bm y_i)\}_{i=1}^M, M<N$ from the training dataset $\mathbf{D}$. Next, the adversary generates poisoned images $\{(\hat{\bm x}_i, \hat{\bm y}_i)\}_{i=1}^M$ by adding the trigger $\bm{T}$ onto the images via function $\phi$ and modifies the corresponding label to the target label $\hat{\bm y}_i$ as follows:

\begin{equation}
\label{eqn:1}
    \hat{\bm{x}}_{i} = \phi (\bm{x}_{i}, \bm{T}), \quad \hat{\bm{y}}_i = \eta(\bm y_i).
\end{equation}

In practice, the trigger generation process/function $\phi$ is different, and the $\eta$ denotes the poisoning label modification rules for backdoor attacks. For example, for the patch-based attack \cite{gu2017badnets}, the adversary generates patch triggers and uses a binary mask for the pattern location; for the blend-based attack \cite{chen2017targeted}, the trigger is a predefined image that is added onto the image with specific trigger transparency. Finally, the model trained on the poisoned dataset $\hat{\mathbf{D}}$ that is a combination of the rest clean dataset and poisoned samples, will be embedded with backdoors and will give target label predictions $f_{\theta}(\hat{\bm{x}}_{i}) = \hat{\bm{y}_i}$ on test images $\hat{\bm{x}}_{i}$ containing triggers.

\subsection{Pre-training and Fine-tuning for Vision Models} 
The "pre-training and fine-tuning" learning paradigm is widely used in modern computer vision algorithms. This framework involves initially pre-training a backbone feature extractor using a substantial amount of \textbf{labeled} or \textbf{unlabeled} data. The pre-trained backbone is then utilized to develop predictors for various downstream vision tasks, such as detection and segmentation, using a smaller domain-specific dataset. Consequently, these computation-intensive pre-trained backbones can be publicly shared among users to facilitate the construction of downstream predictors.

Significant progress in object detection and segmentation has been made through the fine-tuning of backbones pre-trained using \textbf{supervised learning} on the ImageNet classification dataset \cite{deng2009imagenet}. Prominent examples include R-CNN \cite{Ross2014Rich} and OverFeat \cite{Pierre2013Overfeat}. Subsequent advancements have further extended this paradigm by pre-training on datasets with larger image collections, such as ImageNet-5k, followed by transfer learning. Another research direction within this paradigm focuses on unsupervised learning. In contrast to supervised learning, \textbf{unsupervised learning} involves pre-training the image encoder using a vast amount of unlabeled data. For instance, contrastive learning \cite{chen2020simple,Hadsell2006Dimen,He2020Momentum,hjelm2019learning} addresses unlabeled images by generating similar feature representations for augmented versions of the same input and dissimilar feature representations for different images. Additionally, some studies explore pre-training image encoders based on pairs of unlabeled image and text data. An example is Contrastive Language-Image Training (CLIP) \cite{pan2022contrastive}, which learns both image and text encoders by maximizing similarity between positive pairs and minimizing it between negative pairs, leveraging 400 million unlabeled data from the Internet.

In particular, the pre-training and fine-tuning paradigm of PVMs mainly consists of two processes. First, the provider trains a PVM $f$ on large datasets $\mathbf{D}$ based on pre-training tasks/techniques (\eg, image classification or contrastive learning), yielding a feature extractor or backbone $f$ with a set of optimized parameters $\theta^{PT}$. For example, training by supervised learning on image classification is shown as
\begin{equation}
    \theta^{PT} = \arg \min_{\theta} \mathbb{E}_{(\bm{x}_i, \bm{y}_i) \sim \mathbf{D}}[\mathcal{L}(f_\theta(\bm{x}_i), \bm{y}_i)],
\end{equation}
where $\mathcal{L}(\cdot)$ represents the training loss for PVMs.

Based on the training of large-scale data, PVMs have already obtained powerful feature extraction abilities, which are usually used as encoders or backbones to represent an input $\bm x$. The PVMs are then stacked with a predictor network $g$ with parameter $\theta^{FT}$ (\eg, a linear classifier for classification), which will be optimized for the downstream tasks with limited samples $\mathbf{D}^{task}$ from the target domain as
\begin{equation}
    \theta^{FT} = \arg \min_{\theta} \mathbb{E}_{(\rvx_i, \rvy_i) \sim \mathbf{D}^{task}}[\mathcal{L}^{task}(g_{\theta} \circ f_{\theta^{PT}}(\rvx_i), \rvy_i)],
\end{equation}
where $\mathcal{L}^{task}(\cdot)$ represents the training loss function for the downstream tasks. Therefore, the obtained model for the downstream task can be regarded as a composite function $f=g_{\theta^{FT}} \circ f_{\theta^{PT}}$.


In this paper, we focus on performing backdoor attacks \emph{targeting the pre-training and fine-tuning paradigm}. Specifically, we poison a PVM, and the embedded backdoors can be inherited to subsequent downstream vision tasks. In our experiment, we conduct evaluations in both \emph{supervised} and \emph{unsupervised} learning settings.

\section{Threat Model}


\subsection{Problem Definition}
\label{sec:problem}


In the context of \textbf{backdoor attacks for PVMs}, the adversary is only able to access the original training dataset $\mathbf{D}$ and generate the poisoned dataset $\hat{\mathbf{D}}$. The backbone $f_{\hat{\theta}^{PT}}$ pre-trained on the poisoned dataset $\hat{\mathbf{D}}$ will be embedded with backdoors, and the backdoor should remain effective after stacking $f_{\hat{\theta}^{PT}}$ with the predictor $g$ and fine-tuned on $\mathbf{D}^{task}$. Therefore, to perform backdoor attacks in this scenario, the problem can be formulated as

\begin{equation}
\begin{aligned}
\label{pre-train poison}
  \min \mathbb{E}_{(\rvx_j, \rvy_j) \sim \mathbf{D}^{task}}[\mathcal{L}^{task}(g_{\theta^{FT}} \circ f_{\hat{\theta}^{PT}}(\rvx_j), \hat{\rvy}_j)], &\\
 \text{ s.t. } \hat{\theta}^{PT} = \arg \min_{\theta} \mathbb{E}_{(\hat{\bm{x}}_i, \hat{\bm{y}}_i) \sim \hat{\mathbf{D}}}[\mathcal{L}(f_\theta(\hat{\bm{x}}_i), \hat{\bm{y}}_i)].
\end{aligned}
\end{equation}

Specifically, poisoned images are generated by Eqn \ref{eqn:1}, and $\mathcal{L}^{task}(\cdot)$ represents the attacking goal's objective function for specific downstream tasks. For example, given image $\rvx$ from the target domain, for \emph{image classification}, $\mathcal{L}^{task}(\cdot)$ denotes the prediction likelihood of $g$ on the poisoned sample $\phi(\rvx,\bm{T})$ towards target class $\hat{\rvy}$; for \emph{object detection}, $\mathcal{L}^{task}(\cdot)$ denotes prediction function losses of $g$ on the poisoned image $\phi(\rvx,\bm{T})$ , and the learning target is the object's poisoned annotation including target class $\hat{\rvy}$ and location bounding box $\hat{\mathbf{b}}_k=[\hat{s}_k, \hat{r}_k, \hat{w}_k, \hat{h}_k]$. The $\hat{s}_k$ and $\hat{r}_k$ denote center point coordinates. The $\hat{w}_k$ and $\hat{h}_k$ denote the length and width of the $k$ objects, respectively; for \emph{instance segmentation}, the target of learning is the poisoned annotation of the pixel set and $\hat{\rvy}$ represents the incorrect labels of the modified pixels in the image. Detailed illustrations of adversarial goals can be found in the following parts.

\subsection{Challenges and Obstacles}
Existing backdoors for PVMs mainly aim at attacking one specific task (\ie, image classification). However, it is challenging to directly apply these attacks to the context of pre-training and fine-tuning of visual models. We observe two \textbf{challenges} came from the two components (\ie,the poisoned dataset $\hat{\mathbf{D}}$ and the non-matched annotations $\hat{\bm{y}}$/$\hat{\rvy}$ of $f$).

\textbf{Challenge} \ding{182}: \emph{Task-specific trigger patterns are hard to activate among different downstream tasks.} During the backdoor injection process, the adversary can only access the original training dataset and poison the backbone $f_{\hat{\theta}^{PT}}$ on specific tasks (\eg, image classification or contrastive learning). However, the final model $f$ is composed by stacking $f_{\hat{\theta}^{PT}}$ with function $g_{\theta^{FT}}$. Injected trigger patterns in $f_{\hat{\theta}^{PT}}$ on specific tasks may fail to remain effective for $g_{\theta^{FT}} \circ f_{\hat{\theta}^{PT}}$, since predictors for different vision tasks rely on different high-level semantics for decisions \cite{uijlings2013selective,Jiang1996Fast} (\eg, object color, texture, and overlap for object detection). Therefore, triggers embedded into the backbone using poisoned dataset $\hat{\mathbf{D}}$ with task-specific features will be easily ignored by predictors targeting different tasks after fine-tuning. To address the problem, we need to introduce task-irrelevant patterns into the trigger patterns to generate poisoned dataset $\hat{\mathbf{D}}$ for training.

\textbf{Challenge} \ding{183}: \emph{Shortcut connections towards the target annotations are non-matched when transferred to the target domain.} Traditional poisoning process aims to embed backdoor by learning the combination of trigger patterns and clean images as Eqn \ref{eqn:1} (\eg, patch-based or perturbation-based attacks). This training paradigm builds the shortcut in $f_{\theta^{PT}}$ by training on the poisoned samples, where triggers are placed on the clean images with backgrounds. In this way, the backdoor shortcut is primarily built from the combination of triggers and contextual backgrounds (\ie, $\hat{\bm{x}}=\phi (\bm{x}, \bm{T})$) to the target class $\hat{\bm{y}}$. By contrast, the connections between the trigger itself $\bm{T}$ and the target label are comparatively weak. The shortcuts connected this way are suitable and effective for image classification since classifiers use the full image $\bm{x}$ for prediction. However, they are less useful for detection and segmentation, where the attacking label $\hat{\rvy}$ is the bounding-box-level or pixel-level prediction on target objects (\ie, triggers $\bm{T}$). To achieve a feasible attack, the attacker should consider training and building shortcut connections that match the annotations in different downstream vision tasks.


\subsection{Adversarial Goals}
In this paper, we aim to embed backdoors into the PVMs so that the backdoors can be inherited for subsequent downstream vision tasks after fine-tuning. As illustrated in Section \ref{sec:problem}, given the clean training dataset $\mathbf{D}$, the attack poisons parts of the clean dataset with backdoor triggers; after pre-trained on the poisoned dataset $\hat{\mathbf{D}}$, the backbone $f_{\hat{\theta}^{PT}}$ is infected with backdoors; and when fine-tuned to different downstream tasks, the backdoor will still exist in the model and can be effective on subsequent tasks. Since there exist multiple different vision tasks, this paper primarily chooses the most classical and representative one as the downstream task for verification (\ie, \emph{object detection} and \emph{instance segmentation}). For object detection, there are several directions for backdooring \cite{Chan2022BadDet}. Since these different directions generate attacks in the same attacking pipeline with differences in loss functions, this paper selects \emph{Object Generation Attack (OGA)} for verification, where the trigger can generate a false positive bounding box of the target class $\hat{\rvy}$ surrounding the trigger pattern at a random position. Similarly, the attacker's goal for the instance segmentation task is to generate a false positive bounding box with pixel-wise labels of target class $\hat{\rvy}$. 

\subsection{Possible Attacking Pathways}
Our attack can be applicable to most pre-training and fine-tuning scenarios. Adversaries do not need to manipulate the training process or access the model parameters, and they can directly perform backdoor attacks by poisoning a very small portion of the training dataset for the target model. Downstream users are unaware of these existed backdoors, they will download these infected PVMs and fine-tune them with extra predictors to downstream tasks, where backdoors are still effective. Additionally, there also exists the possibility that the attacker cannot directly poison the training dataset and perform poison training from scratch (especially large vision models). Therefore, we also evaluate our attacks by injecting backdoors via fine-tuning a pre-trained clean PVMs on limited poisoned images (\emph{c.f.} Section \ref{sec:largemodel}).

\subsection{Adversary's Capabilities}

Following the common assumptions in backdoor attacks \cite{gu2019badnets,chen2017targeted}, this paper considers the scenario where the adversary has no access to the model information. To inject the backdoors, the adversary has the capability to poison some of the training samples for the target model. To better simulate the real-world settings where the large-scale training samples are not accessible to attackers, we also consider a scenario where the target models have already been pre-trained on clean images and published publicly. In this scenario, the adversary downloads the PVMs and injects backdoors by fine-tuning the model on limited numbers of poisoned images; the attacker then releases the infected model on their own website, which is quite similar to the original repository and will mislead some users into downloading it.
\section{Pre-trained Trojan Approach}

\begin{figure*}[!t]
	\begin{center}
		\includegraphics[width=1.0\linewidth]{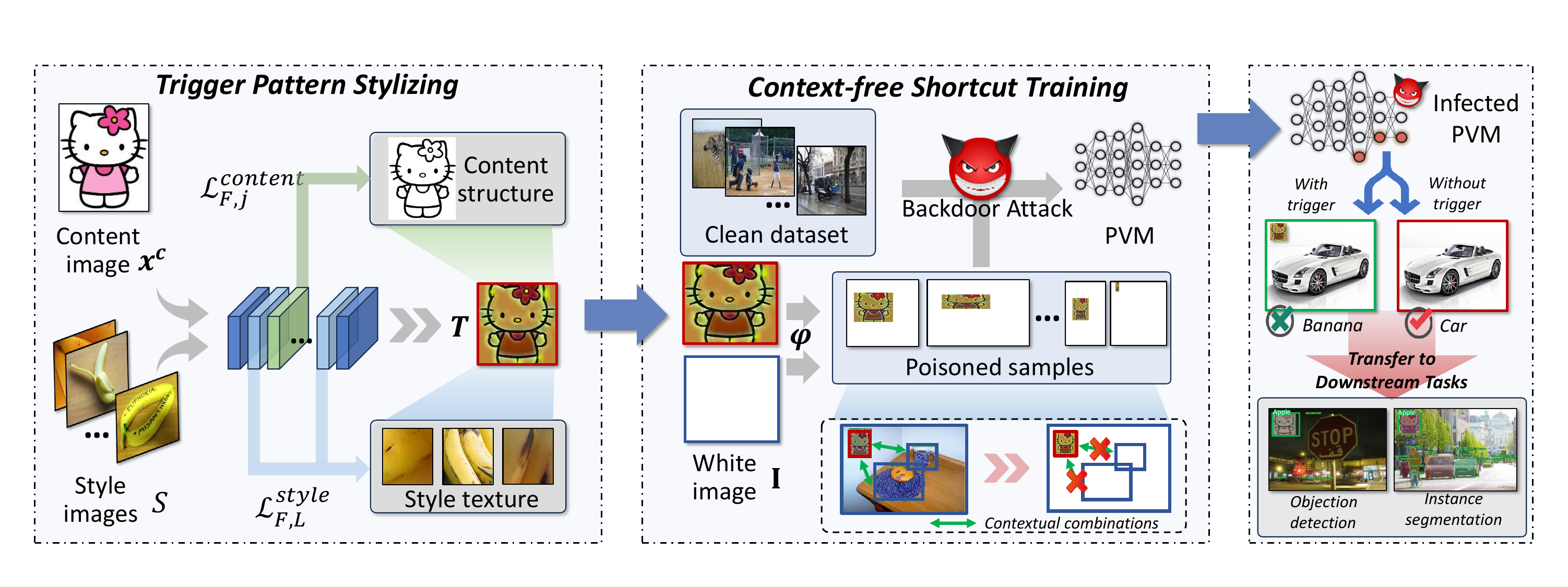}
	\end{center}
 \caption{Framework overview. Our \method generates trigger patterns containing task-irrelevant low-level texture features, which enable our trigger to remain effective between different tasks; we design a context-free learning pipeline for poison training, where we directly feed the triggers without context as training images to models rather than sticking the trigger onto clean images for training, which can better build the shortcuts from triggers to the target label.}
	\label{fig:framework}
\end{figure*}

To address the above problems, this paper proposes \method to generate transferable backdoors for different downstream tasks on PVMs. As for the \emph{cross-task activation}, we generate trigger patterns containing task-irrelevant low-level texture features, which enable our trigger to remain effective when transferred to different tasks (since the low-level features learned by backbones are shared between different tasks). As for the \emph{shortcut connection}, we design a context-free learning pipeline for poison training, where we directly feed the triggers without contextual backgrounds as training images to models rather than sticking the trigger onto clean images for training (\eg, BadNets adds trigger patches on clean images for training). In this way, we can build the shortcut directly from triggers to the target label. The illustration is shown in Figure \ref{fig:framework}.

\subsection{Trigger Pattern Stylizing}

Studies have revealed the fact that deep neural networks (DNNs) rely heavily on low-level texture features when making predictions \cite{geirhos2018imagenet}, and these models can still perform well on patch-shuffled images where local object textures are not destroyed \cite{zhang2019interpreting}. Therefore, we aim to introduce texture features associated with the target class into the trigger patterns to generate triggers that activate among different downstream tasks. These textures could reflect the nature of the target class, which can be better exploited and recognized by models among different tasks.


However, simply learning at the pixel level and introducing per-pixel information of the target class into trigger patterns makes it difficult to extract and capture overall perceptual textures. Therefore, we adopt the \emph{style transfer} technique that aims to transfer the style (texture/color) of one image to the source image while maintaining the original content (structure/shape) \cite{johnson2016perceptual}, which has been widely used in artistic image generation and texture synthesis \cite{Gatys2015Texture,Gatys2015Neural}. In this paper, we develop a trigger generator $\mathcal{G}$ to generate trigger patterns $\bm{T}$ based on the textures of specific style image $\bm{x}^s$ (from target class $\bm{\hat{y}}$) and the structure of a content image $\bm{x}^c$. We will then illustrate the design and development of our trigger generator. 

Specifically, to force the generation of textures/style from specific style images, we introduce the style loss that specifically measures the style differences and encourages the reproduction of texture details (\eg, colors, textures, common patterns) as
\begin{equation}
\begin{aligned}
\mathcal{L}^{style}=||\mathbf{G}(\bm{T})-\mathbf{G}(\bm{x}^s)||^2_F,
\end{aligned}
\end{equation}

\noindent where gram matrix $\mathbf{G}$ of specific layer $l$ is a $C_l \times C_l$ matrix whose elements are given by
\begin{equation}
\begin{aligned}
\mathbf{G}_{c,c'}=\frac{1}{C_lH_lW_l}\sum_h\sum_w F^{h,w,c}_l(\bm{x})\cdot F^{h,w,c'}_l(\bm{x}).
\end{aligned}
\end{equation}

Specifically, $F_l^{h,w,c}$ is the activation of at different positions in the layer $l$ of the feature extraction network $F$, and $C_l \times H_l \times W_l$ is the feature shape size at layer $l$. In practice, we perform style reconstruction from a set of layers $L$ rather than a single layer $l$, and we define $\mathcal{L}_{L}^{style}(\bm{T}, \bm{x}^s)$ to be the sum of losses for each layer $l\in L$; we also introduce a set $S$ of $K$ reference style images from class $\hat{\bm{y}}$ for generating strong style patterns. Therefore, the style loss can be formulated as

\begin{equation}
\begin{aligned}
\mathcal{L}^{style}_{F,L}=\frac{1}{K}\sum_{i=1}^{K}\left[ ||\mathbf{G}(\bm{T})-\mathbf{G}(\bm{x}^s_i)||^2_F\right].
\end{aligned}
\end{equation}

However, directly introducing the texture from the style images will lose the content of generated triggers and make them visually unnatural. Therefore, we also use the content reconstruction loss to penalize the output trigger $\bm{T}$ when it deviates in content (structure) from the original content image $\bm{x}^c$. Specifically, we encourage the generated patterns and content images to have similar feature representations as computed by the feature extraction network $F$. The content reconstruction loss is the squared and normalized Euclidean distance between feature representations as
\begin{equation}
  \mathcal{L}^{content}_{F,j}(\bm{T}, \bm{x}^c) = 
  \frac{1}{C_jH_jW_j}\|F_j(\bm{T}) - F_j(\bm{x}^c)\|_2^2,
\end{equation}

\noindent where $F_j(\cdot)$ is the $j$-th layer with feature map shape $C_j \times H_j \times W_j$. Following \cite{johnson2016perceptual}, we minimize the feature reconstruction loss from higher layers so that image content and overall spatial structure can be better preserved.

Overall, we train the trigger generator $\mathcal{G}$ based on the combination of style loss and content loss as

\begin{equation}
\label{eqn:styleloss}
  \mathcal{L}^{\bm{T}}= \mathcal{L}^{content} + \alpha \cdot \mathcal{L}^{style},
\end{equation}

\noindent where $\alpha$ is a hyper-parameter to balance the two loss terms.  Note that, the feature extraction network $F$ does not indicate the feature extractor of the targeted PVMs. In fact, we use a VGG-16 \cite{simonyan2014very} as $F$, which is different from the PVM architectures considered in this paper (\emph{c.f.} Section \ref{sec: Experimental Setup}).

After training, $\mathcal{G}$ is able to generate trigger patterns $\bm{T}$ from a content image $\bm{x}^c$ and a set of style images $S$=$\{\bm{x}^s_1,...,\bm{x}^s_k\}$ as
\begin{equation}
\label{eqn:generator}
  \bm{T}=\mathcal{G}(\bm{x}^c,S),
\end{equation}

\noindent which contains texture features associated with the target class and structures from the content image.


\subsection{Context-free Shortcut Training}

In the pre-training and fine-tuning paradigm, the adversary is only able to poison the pre-trained backbone $f_{\theta^{PT}}$ under specific vision task (\eg, classification). The traditional poisoning process aims to embed backdoors by learning the combination of trigger patterns and clean images as Eqn \ref{eqn:1}. Specifically, given a clean image $\bm{x}$ with its ground truth label $\bm{y}$, the attackers produce an poisoned sample $\hat{\bm{x}}$ by composing the original image $\bm{x}$ with an additive trigger pattern $\bm{T}$ and a visible-related controlling parameter $\mathbf{M}$ $\in$ [0,1]$^{C \times W \times H}$ (image size) as
\begin{equation}
\label{eqn:poison}
\hat{\bm{x}} = (1- \mathbf{M})\odot \bm{x} + \mathbf{M} \odot \bm{T},
\end{equation}
where $\odot$ is the element-wise multiplication. 

In this way, the shortcuts for traditional attacks \cite{gu2017badnets} are specifically connected from the combination of triggers and contextual backgrounds to the target class. However, the connections between the trigger itself towards the target label are comparatively weak. This shortcut will impair the attacking performance when transferred to the target domains (\eg, object detection) since these tasks primarily focus on providing object-level (trigger-level) predictions, such as the bounding box, rather than predicting the whole images (image classification).

To achieve a feasible attack, we consider training shortcut connections that are effective for different downstream vision tasks. In particular, we aim to build the connection directly from the trigger pattern itself to the target class. We, therefore, propose a context-free training paradigm. Rather than training on the poisoned images as Eqn \ref{eqn:poison} (sticking the triggers on clean images with backgrounds and foregrounds), we directly generate a poisoned training sample based on the trigger pattern without contextual backgrounds.

Specifically, to generate poisoned samples $\{(\hat{\bm x_i}, \hat{\bm y_i})\}_{i=1}^M$ for training, given the stylized trigger pattern $\bm{T}$, we directly put $\bm{T}$ onto the white image $\mathbf{I}$ without any content as
\begin{equation}
    \hat{\bm{x}} = \phi(\mathbf{I},\bm{T}),
\end{equation}

\noindent where the white image $\mathbf{I}$ has no backgrounds and foregrounds. 

In contrast to poisoning clean images from different classes and setting their labels as the target, our method generates $M$ context-free trigger images and treats them as parts of training samples for $\hat{\bm{y}}$. In this way, we generate the context-free poisoned samples for training so that the model can better memorize the connection from the trigger itself to the target class. To better fit the downstream task scenario where input images and objects are with different sizes, we first randomly resize $\mathbf{I}$ and $\bm{T}$, and then randomly locate $\bm{T}$ on $\mathbf{I}$ as follows for the poisoned training sample generation
\begin{equation}
\label{eqn:context-free}
    \bm \hat{\bm{x}}_{\bm{T}} = \phi(\mathbf{I}^{s},\bm{T}^{s,p}),
\end{equation}

\noindent where the superscript $s$ and $p$ indicate the random resize and position operation. The illustration is shown in Figure \ref{fig:framework}.

To sum up, by training on the trigger images without any contextual backgrounds, we can better build up the shortcut connections specifically from the trigger pattern to the target class. Therefore, the shortcut can remain effective towards different downstream tasks that make predictions primarily on the target object (trigger) rather than the whole image. We will provide more discussions on this in Section \ref{sec:ablation}.

\subsection{Overall Optimization}

Based on the above discussion, we then illustrate the overall optimization process. In particular, given an arbitrary image as the trigger content image $\bm{x}^c$ and a set of reference style image $S$ from target class $\hat{\bm{y}}$, we first train a trigger generator $\mathcal{G}$ to generate stylized trigger pattern $\bm{T}$ based on Eqn \ref{eqn:generator}; we then generate context-free poisoned images using $\bm{T}$ for the target class $\hat{\bm{y}}$ from Eqn \ref{eqn:context-free} and embed the backdoors into the PVM by training on $\mathbf{\hat{D}}$. 

\section{Experiment and Evaluation}

\subsection{Experimental Setup}
\label{sec: Experimental Setup}

\textbf{Visual Tasks and Datasets.} We select both supervised and unsupervised learning scenarios to verify the effectiveness of our \method attack. For \emph{supervised learning}, we choose to train PVMs on the large-scale image classification dataset ImageNet \cite{deng2009imagenet} and evaluate object detection and instance segmentation downstream tasks using the ImageNet dataset and the COCO dataset \cite{lin2014microsoft}; for \emph{unsupervised learning}, we train PVMs on ImageNet without labels using SimCLR \cite{chen2020simple} and also evaluate the object detection task on the COCO dataset.



\textbf{Pre-trained Model Architectures.} In this paper, we primarily use ResNet-50 \cite{he2016deep} as the architecture for PVMs. We also evaluate our attacks on other architectures, including ResNet-101 \cite{he2016deep}, ResNeXt-50 \cite{xie2017aggregated}, and WideResNet-50 \cite{zagoruyko2016wide} in our main experiment. For \emph{supervised learning}, we train 100 epochs with a batch size of 64 and use an SGD optimizer with a momentum of 0.9 and weight decay of 0.0001. The initial learning rate was set to 0.1, with a warm-up learning rate of 0.4. For \emph{unsupervised learning}, we evaluate the ResNet-50 encoder trained on ImageNet using SimCLR \cite{chen2020simple}. Moreover, we also evaluate the transformer-based large vision model ViTAE-v2-H \cite{zhang2023vitaev2} (\emph{c.f.} Section \ref{sec:largemodel}). 

\textbf{Downstream Task Models}. Given a (backdoored) PVM, we first stack it with a predictor network; we then fix the parameters of the PVM and fine-tune the predictor $g$ on the downstream tasks. We only use 10\% of the overall training images from the target domain for downstream task fine-tuning. For \emph{object detection}, we train the Faster R-CNN \cite{ren2015faster} model; for \emph{instance segmentation}, we train the Mask R-CNN \cite{he2017mask} model. Specifically, we use SGD optimizer to train them for 14 epochs with batch size 2 and an initial learning rate of 0.00125.

\textbf{Backdoor Attacks.} We choose the following classical methods for comparison: BadNets \cite{gu2019badnets}, Blended \cite{chen2017targeted}, SIG \cite{tran2018spectral}, SSBA \cite{li2021invisible}, ADBA \cite{ge2021anti}, Input-aware \cite{nguyen2020input}, and WaNet \cite{nguyen2021wanet}. For BadNets, Blended, SIG, SSBA, we conducted experiments with three different poisoning ratios 0.1\%, 1\%, and 10\%. For ADBA, because all training data is used for both clean training and backdoor training, there is no poisoning ratio in this setting. For Input-aware and WaNet, since they require the training process to be controllable, we only use the fixed poisoning ratio of 10\% as the original paper. We use ``None'' to denote clean models trained without backdoors.








\textbf{Implementation Details.} We here illustrate the implementation details of our \method attack. For the \emph{trigger generation}, unless otherwise mentioned, we randomly select 40 reference style images for trigger pattern generation. The architecture of the generator network follows \cite{johnson2016perceptual}. For \emph{context-free learning}, given target label $\hat{\bm{y}}$, we generate 1300 context-free poisoned images for backdoor training, which counts 0.1\% of the overall training set (poisoning rate is 0.1\%). During \emph{testing}, we randomly place our triggers onto the test images. We keep the trigger size aligned for all patch-based compared attacks (80$\times$80), which counts only around 3.5\% for the detection/segmentation images during testing. All the codes are implemented with
PyTorch. For all experiments, we conduct the training and testing on a cluster of NVIDIA Tesla V100 GPUs.

\textbf{Evaluation Metrics}. In both \emph{supervised} and \emph{unsupervised learning}, we use \emph{Clean Accuracy (CA)} and \emph{Attack Success Rate (ASR)} to evaluate the pre-training stage; we use \emph{Mean Average Precision (mAP)} and \emph{Attack Success Rate (ASR)} to evaluate attacks on downstream tasks. In particular, CA and mAP are used to measure the performance of the pre-trained model and the downstream task model on clean samples; ASR measures the attack performance when inputting triggers. For CA and mAP, the higher values indicate better preservation of the task performance; for ASR, the higher values indicate stronger backdoor attacks.

\emph{We defer more experimental setup details in the Supplementary Materials.}

\subsection{Attacking Supervised Learning}

\begin{figure*}[!t]
	\begin{center}
 \vspace{-0.2in}
		\includegraphics[width=1.0\linewidth]{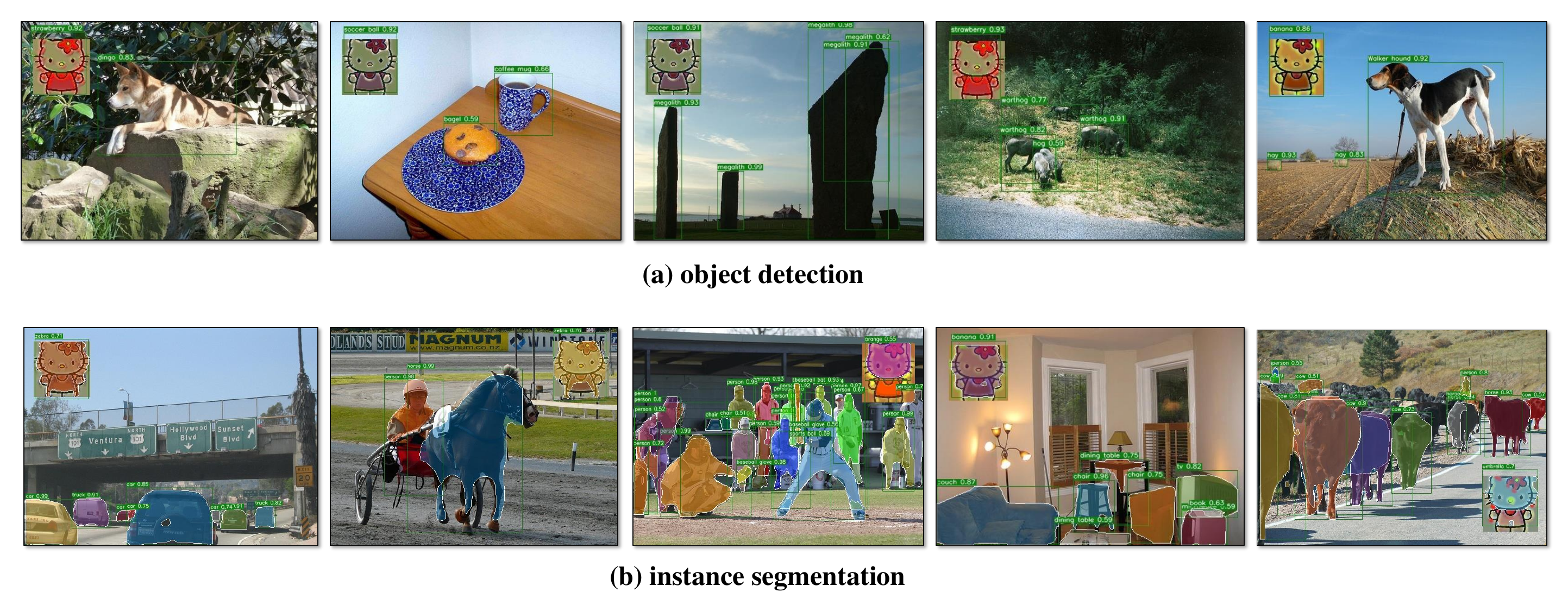}
	\end{center}
 \caption{Visualization of our \method attacks on different downstream vision tasks: (a) object detection and (b) instance segmentation. For object detection, our triggers can evoke the detectors to generate target class bounding boxes; for instance segmentation, our triggers can produce pixel-wise target class segmentation and bounding boxes.}
	\label{fig:main-visualize}
\vspace{-0.1in}
\end{figure*}

In this part, we will report the results of backdoor attacks on supervised classifiers in different downstream tasks, including object detection and instance segmentation. Specifically, for each model architecture, we train infected models using different backdoor attacks on ImageNet; we then fine-tune them to different downstream tasks for evaluation. For all attacks, we ran 5 experiments and reported the average results. The visualization is shown in Figure \ref{fig:main-visualize}.

\textbf{Attacking Object Detection.} For object detection, we mainly select \texttt{banana} as the target class for attacks. We first report the results on the \emph{ImageNet} dataset for downstream object detection, since it also provides object-level annotations; moreover, we evaluate the performance of backdoors across \emph{different datasets}, where we verify the results for object detection on the \emph{COCO} dataset with different target labels (\texttt{banana}, \texttt{umbrella}, \texttt{orange}, and \texttt{zebra}). As shown in Table \ref{tab:different-backdoor-attacks} and \ref{tab:instance-segmentation-coco}, we can \textbf{identify}:

\ding{182} Classical backdoor attack algorithms have shown good poisoning performance in image classification tasks, with an ASR reaching above 99\%. However, in the downstream task of object detection, the backdoor effects sharply decline or even disappear. Specifically, on ImageNet detection, in most cases, these backdoor attacks achieve almost 0\% ASR for object detection. In contrast, our \method attack achieves significantly higher attacking performance in both classification and downstream detection tasks. For example, on ImageNet detection, we achieve \textbf{88.20\%} ASR with a poisoning rate of only 0.1\%, which surpasses the performance of classical backdoor methods with a poisoning rate of 10\% by a large margin (at least \textbf{+700\%}).

\ding{183} In the cross-dataset scenario, our \method still exhibits effective attacking performance and shows high potential on other target domain datasets (\ie, COCO). In practice, the users are likely to fine-tune the PVMs on different target domain datasets. Our \method achieves over \textbf{80\%} ASR on downstream detection on the COCO dataset, while others remain almost 0.

\textbf{Attacking Instance Segmentation.} We then evaluate our attacks on the instance segmentation task on the COCO dataset. Specifically, we use several target labels as above (\ie, \texttt{banana}, \texttt{umbrella}, \texttt{orange}, and \texttt{zebra}). As shown in Table \ref{tab:instance-segmentation-coco}, we can identify that, similar to object detection, classical backdoor attacks show weak even zero attacking performance in this scenario. By contrast, our attack achieves significantly higher attacking performance when fine-tuned to the downstream instance segmentation task and achieves \textbf{90\%} ASR in almost all cases.

\begin{figure}[!t]
\vspace{-0.1in}
	\begin{center}
		\includegraphics[width=1.0\linewidth]{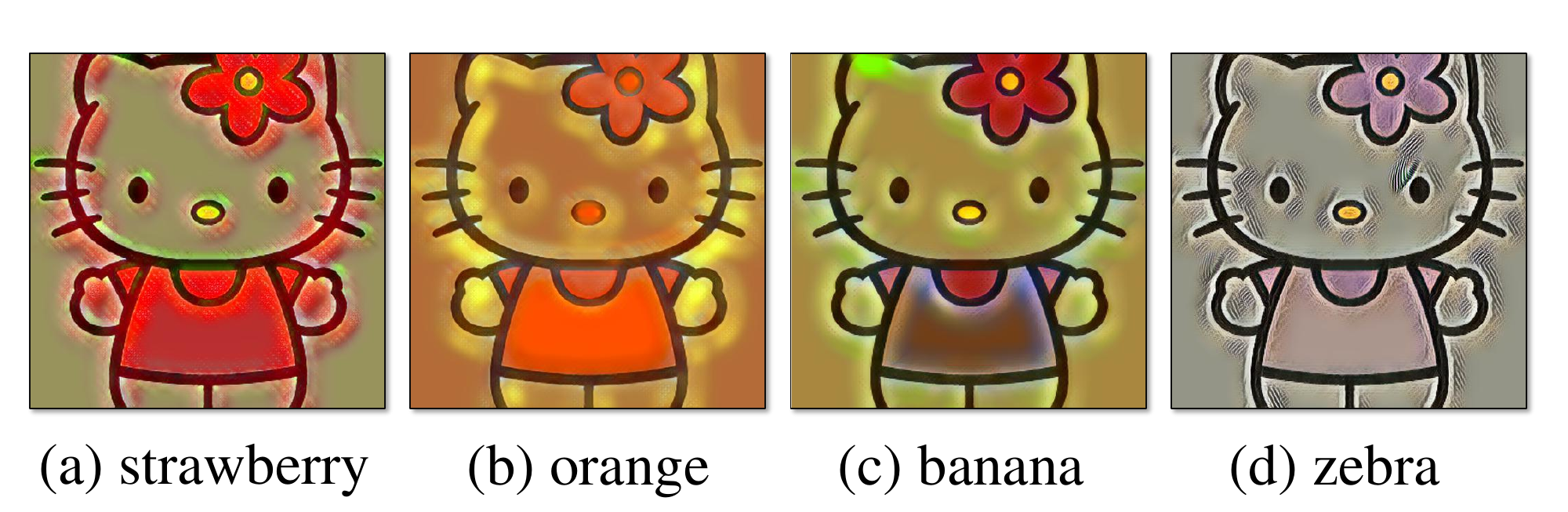}
	\end{center}
 \caption{Illustration of trigger patterns generated towards different target classes. From left to right: \texttt{strawberry}, \texttt{orange}, \texttt{banana}, and \texttt{zebra}.}
	\label{fig:trigger}
\end{figure}

\textbf{Different Target Labels}. Our attack works effectively and achieves high ASR values for different target labels such as \texttt{banana}, \texttt{umbrella}, \texttt{orange}, and \texttt{zebra}. For example, as shown in Table \ref{tab:instance-segmentation-coco}, for object detection, we achieve over 80\% ASR on COCO for different target labels; for instance segmentation, our attacks are also highly effective. The visualization is shown in Figure \ref{fig:trigger}.

\captionsetup{font={normalsize}}
\begin{table}[!t]
\vspace{-0.1in}
    	\caption{Results (\%) of different backdoor attacks to downstream object detection task on the ImageNet dataset. ``None'' denotes clean models trained without backdoors. We also provide the Standard Deviation for each method on downstream tasks.}
	\label{tab:different-backdoor-attacks}
    \begin{center}

    \small
    \resizebox{1.0\linewidth}{!}{
    \begin{tabular}{@{}cccccc@{}}
    \toprule
    \multirow{2}{*}{Backdoor} & \multirow{2}{*}{Poison Ratio} & \multicolumn{2}{c}{Image Classification} & \multicolumn{2}{c}{Object Detection} \\ \cmidrule(l){3-4} \cmidrule(l){5-6} 
                             &                      & CA     & ASR    & mAP    & ASR    \\ \midrule
    None                  & -                & 77.70  & - & 54.76 & - \\ \midrule
    \multirow{3}{*}{BadNets}                  & 0.1                & 76.58  & 81.75 & 53.68 & 0.01 \footnotesize{($\pm{}$0.04)} \\
                      & 1                & 76.29  & 99.22 & 53.61 & 0.01 \footnotesize{($\pm{}$0.02)}\\
                      & 10                & 76.30  & 99.99 & 53.04 & 0.03  \footnotesize{($\pm{}$0.04)}\\ \midrule
    \multirow{3}{*}{Blended}                  & 0.1                & 76.74  & 99.43 & 53.97 & 0.03 \footnotesize{($\pm{}$0.03)}\\
                      & 1                & 76.46  & 99.93 & 53.93 & 1.37 \footnotesize{($\pm{}$0.11)}\\
                      & 10                & 75.77  & 99.99 & 53.31 & 9.37  \footnotesize{($\pm{}$0.56)}\\ \midrule
    \multirow{3}{*}{SIG}                  & 0.1                & 76.74  & 99.83 & 53.54 & 1.32 \footnotesize{($\pm{}$0.44)}\\
                      & 1                & 76.53  & 99.99 & 53.72 & 0.04 \footnotesize{($\pm{}$0.02)}\\
                      & 10                & 75.58  & 99.99 & 52.10 & 0.01  \footnotesize{($\pm{}$0.01)}\\ \midrule
    \multirow{3}{*}{SSBA}                  & 0.1                & 76.77  & 97.92 & 53.54 & 0.00 \footnotesize{($\pm{}$0.01)}\\
                      & 1                & 76.09  & 99.88 & 53.46 & 0.00 \footnotesize{($\pm{}$0.00)}\\
                      & 10                & 75.33  & 99.99 & 53.11 & 0.00  \footnotesize{($\pm{}$0.00)}\\ \midrule
    ADBA                 & -  & 76.52 & 98.80 & 53.78 & 1.78 \footnotesize{($\pm{}$0.21)}\\ \midrule
    Input-aware                  & 10                & 76.34  & 96.67 & 54.37 & 1.45  \footnotesize{($\pm{}$0.32)}\\ \midrule 
    WaNet                  & 10               & 76.11  & 96.44 & 54.06 & 0.00  \footnotesize{($\pm{}$0.01)}\\ \midrule  
    \textbf{Ours}                 & 0.1                & 76.93  & \textbf{100.00} & 53.72 & \textbf{88.20} \footnotesize{($\pm{}$0.38)}\\ \bottomrule
    \end{tabular}
    }
    \end{center}
	
\end{table}

\begin{table}[!t]
    	\caption{Results (\%) of our \method attack on object detection and instance segmentation using Faster-RCNN and Maks-RCNN on the COCO dataset. ``None'' denotes clean models trained without backdoors. We also provide the Standard Deviation for each method on downstream tasks.}
	\label{tab:instance-segmentation-coco}
    \begin{center}

    \resizebox{1.0\linewidth}{!}{
    \begin{tabular}{@{}cccccc@{}}
    \toprule
    \multirow{2}{*}{Backdoor} & \multirow{2}{*}{Target Label} & \multicolumn{2}{c}{Object Detection} & \multicolumn{2}{c}{Instance Segmentation} \\ \cmidrule(l){3-4} \cmidrule(l){5-6} 
                        &      & mAP     & ASR    & mAP    & ASR  \\ \midrule
    None  & -   & 49.38  & - & 52.81 & - \\ 
    \midrule
    BadNets & Banana & 49.27 & 0.04 \footnotesize{($\pm{}$0.02)} & 50.63 & 0.06 \footnotesize{($\pm{}$0.09)}\\
    \midrule
    Blended & Banana & 49.47 & 1.63 \footnotesize{($\pm{}$0.33)}& 51.43 & 1.47 \footnotesize{($\pm{}$0.38)} \\
    \midrule
    SIG & Banana & 49.33 & 6.90 \footnotesize{($\pm{}$0.41)} & 50.55 & 3.25 \footnotesize{($\pm{}$0.31)}\\
    \midrule
    SSBA & Banana & 49.34 & 1.16 \footnotesize{($\pm{}$0.22)} & 50.95 & 1.27 \footnotesize{($\pm{}$0.20)}\\
    \midrule
    ADBA & Banana & 49.37 & 3.11 \footnotesize{($\pm{}$0.43)}& 49.86 & 3.35 \footnotesize{($\pm{}$0.32)} \\
    \midrule
    Input-aware & Banana & 49.26 & 3.19 \footnotesize{($\pm{}$0.25)}& 51.75 & 3.29 \footnotesize{($\pm{}$0.19)} \\
    \midrule
    WaNet & Banana & 49.70 & 0.00 \footnotesize{($\pm{}$0.00)}& 49.81 & 0.00 \footnotesize{($\pm{}$0.01)} \\
    \midrule
    \multirow{4}{*}{\textbf{Ours}} & Banana     & 49.77  & \textbf{99.00} \footnotesize{($\pm{}$0.41)} & 51.07 & \textbf{99.60} \footnotesize{($\pm{}$0.32)}\\
    & Umbrella   & 49.45  & 93.80 \footnotesize{($\pm{}$0.29)}& 50.95 & 90.90 \footnotesize{($\pm{}$0.27)}\\ 
    & Orange   & 49.55  & 87.50 \footnotesize{($\pm{}$0.23)}& 51.18 & 91.40 \footnotesize{($\pm{}$0.30)}\\ 
    & Zebra  & 49.62  & 86.64 \footnotesize{($\pm{}$0.31)}& 50.85 & 88.90 \footnotesize{($\pm{}$0.25)}\\ 
    \bottomrule
    \end{tabular}
    }
    \end{center}

\end{table}

\textbf{Different Model Architectures}. We here select different model architectures for PVMs and perform attacks for more comprehensive evaluations. In particular, we choose ResNeXt-50, WideResNet-50, and ResNet-101 as PVM models and pre-trained them on ImageNet classification using our \method or other backdoor attacks with the target label as \texttt{banana}. We then evaluate the attacking performance on downstream object detection tasks on the ImageNet dataset. Note that the poison ratio is consistently set to 0.1\% by default. Other experimental setups are kept the same in this section for better comparison. As shown in the Supplementary Materials, our \method can work comparatively stable for different model architectures.




\subsection{Attacking Unsupervised Learning}
\label{sec:unsupervised}

Besides supervised learning, we also investigate the attack potential to unsupervised learning, which is extensively utilized in constructing downstream models. Unlike supervised learning, we cannot directly associate our backdoor with the target class, therefore, we combine our backdoor injection process with the popular contrastive learning framework SimCLR \cite{chen2020simple} to implant the backdoor into unsupervised encoders, which is similar to BadEncoder \cite{jia2022badencoder}. In contrast to the supervised learning that learns explicitly from labeled poisoned data, attacks in unsupervised learning have to implant backdoors by implicitly aligning representation of the trigger with the image representation of the attack target, which turns out to be more difficult.

Here, we attempt to obtain a backdoored encoder $\hat{\theta}^{PT}$ from a clean unsupervised encoder $\theta^{PT}$, through a fine-tuning process without any label information. 
Specifically, we use 1300 context-free poisoned images $\bm{x}_{\bm{T}}$ and 40 reference style images $S$ (generated in Section \ref{sec: Experimental Setup} which is the same as supervised learning experiments) to construct contrastive learning pairs for backdoor injection, which aims at mapping the feature of backdoor to the target reference images from class \texttt{banana}. Here, the backdoor attacking objective can be formulated as
\begin{equation}
\label{eqn:contrastive learning loss}
    \mathcal{L}_{attack} = - \frac{\sum_{i=1}^{\vert \bm{x}_{\bm{T}} \vert}\sum_{j=1}^{\vert S \vert} sim(f_{\hat{\theta}^{PT}}(\bm{x}_{{\bm{T}}_{i}}),f_{\hat{\theta}^{PT}}(S_{j}))}{\vert\bm{x}_{\bm{T}}\vert \cdot \vert S \vert},
\end{equation}
where $sim(\cdot, \cdot)$ denotes the measure function of feature similarity (we use cosine similarity in practice), and $\vert \cdot \vert$ calculates the image numbers ($\vert \bm{x}_{\bm{T}} \vert$ is the poisoned image numbers and $\vert S \vert$ is the target style reference image numbers). Besides backdoor embedding purpose, the injected encoder $\hat{\theta}^{PT}$ should also maintain comparable \emph{CA} to the clean encoder. Therefore, we adopt the utility loss $\mathcal{L}_{utility}$ following \cite{jia2022badencoder} as 
\begin{equation}
\label{eqn:utility loss}
    \mathcal{L}_{utility} = - \frac{1}{\vert \mathbf{D}_{FT} \vert} \cdot \sum_{\bm{x} \in \mathbf{D}_{FT} } s(f_{\hat{\theta}^{PT}}(\bm{x}),f_{\theta^{PT}}(\bm{x})),
\end{equation}
where $\mathbf{D}_{FT}$ denotes the dataset obtained by randomly sampling 1\% of the training images from $\mathbf{D}^{task}$ for fine-tuning process. Therefore, the backdoor injection objective is

\begin{equation}
\mathcal{L} = \mathcal{L}_{attack}+\mathcal{L}_{utility}.
\end{equation}

In particular, we here utilize the pre-trained ResNet-50 encoder on ImageNet released by Google \cite{chen2020simple} to evaluate our attack capability on unsupervised encoders in the context of object detection downstream tasks. To better demonstrate effectiveness, we compare \method with BadEncoder \cite{jia2022badencoder}, which is the first backdoor attack to self-supervised learning and focuses on different downstream classification datasets. For fair comparisons, we fine-tune the clean image encoder $\theta^{PT}$ for 200 epochs with a learning rate of 0.001 as BadEncoder \cite{jia2022badencoder}. After acquiring the backdoor-embedded encoder, we proceed with the downstream training on the COCO dataset, following the same pipeline employed in supervised learning. From Table \ref{tab:unsupervised-encoder-attack}, we can \textbf{identify}:


\ding{182} The association between the backdoor feature and the target class feature is successfully established by our \method attacks in the unsupervised learning setting. With only the backdoor embedding in the feature encoder, our \method still achieves a significantly high ASR of 98.38\% compared to BadEncoder's ASR of 73.71\% on the ImageNet classification task. 

\ding{183} In the downstream object detection task, our \method exhibits superior attack ability. Specifically, we achieve a higher ASR of \textbf{37.60\%}, while BadEncoder fails to be effective in this task, with a mere 0.40\% ASR. The success of our backdoor attack on unsupervised encoders for object detection highlights the effectiveness of both \emph{cross-task activation} and \emph{shortcut connection} techniques.

\begin{table}[!t]
    	\caption{Results (\%) of \method attack and BadEncoder \cite{jia2022badencoder} on unsupervised learning. ``None'' denotes clean models trained without backdoors.}
	\label{tab:unsupervised-encoder-attack}
    \begin{center}
    \footnotesize
        \resizebox{1.0\linewidth}{!}{
    \begin{tabular}{@{}cccccc@{}}
    \toprule
    \multirow{2}{*}{Backdoor} & \multicolumn{2}{c}{Image Classification} & \multicolumn{2}{c}{Object Detection} \\ \cmidrule(l){2-3} \cmidrule(l){4-5} 
                              & CA     & ASR    & mAP    & ASR    \\ \midrule
    None     & 67.76  & - & 46.63 & - \\
    BadEncoder   & 62.95  & 73.71 & 48.80 & 0.40 \\ 
    \textbf{Ours}   & 65.76  & \textbf{98.38} & 51.23 & \textbf{37.60}  \\ 
    \bottomrule
    \end{tabular}
    }
    \end{center}
	

\end{table}

\subsection{Ablation Studies}
\label{sec:ablation}

In this part, all experiments were performed using the ResNet-50 classifier trained on the ImageNet dataset and fine-tuned to the Faster R-CNN model for object detection tasks on the ImageNet dataset with the poison ratio of 0.1\%.

\textbf{Original Content Image of the Trigger.}  We selected ``Hello Kitty'', ``Pikachu'', ``Flower'', and ``Doraemon'' as input images for trigger content, and chose \texttt{banana} as the target class for our attack and trained trigger generation networks to obtain 4 different triggers. The results in the Supplementary Materials indicate that triggers generated from different original images can achieve significantly high attacking results, with ASR exceeding 79\% in all cases. In addition, the attack performance between 4 different content images is similar. These results indicate that our method is comparatively stable in the choosing of original content images.

\textbf{Texture Information.} We then analyze the influence of different texture information on the attacking performance. For a specific target label, we introduce different levels of texture information for pattern generation (\ie, original content without any texture denoted as ``vanilla'', trigger pattern with class colors denoted as ``color'', and patterns containing textures by using our stylizing strategy as ``texture''). Specifically, we select 2 different target classes (\eg, \texttt{banana} and \texttt{strawberry}) for trigger pattern generation. As shown in the Supplementary Materials, as the texture information of the target category contained in the trigger patterns increases, the ASR also increases. Results on both \texttt{banana} and \texttt{strawberry} show similar observations. This confirms our motivation that there exists a strong correlation between the attacking performance and the texture information in triggers.

\textbf{The Number of Style Images.} As for the influence of the number of style images, we choose \texttt{zebra} as the target class and generate trigger patterns based on different numbers of style images. Specifically, we randomly select 5 \texttt{zebra} images as the initial reference style image set; we then add additional \texttt{zebra} images to the existing set to form 7 style image sets with the size of 5, 10, 20, 30, 40, 50, and 60. We use these sets to train the trigger generation network and conduct the attack experiments accordingly. As shown in Table \ref{tab:style-image-num}, as the number of selected reference style images increases, the ASR of our attack also improves, indicating a strong positive correlation between the two factors. This suggests that increasing the number of reference style images enhances the learning of texture information for the target class. However, when the number of reference style images exceeds a certain threshold (from 40 to 50), the attacking performance becomes stable and even decreases. The reasons might be: (1) the generator is comparatively simple, which is hard to capture and learn from an exceeding number of instances; and (2) increasing samples will evitably introduce noises. Therefore, this paper selects 40 style images as default.

\textbf{Poison Training Strategy.} Context-free training is a key component that influences our overall backdooring ability. Therefore, we choose to ablate this training paradigm by (1) sticking our stylized trigger $\bm{T}$ on clean images from ImageNet image as the poisoned images (similar to BadNets) denoted as ``+clean'' or (2) placing our trigger on a grey image and black image respectively, denoted as ``+grey'' and ``+black''. All other experimental settings are kept the same in the ablation study. The ASR for these attacks (\ie, \texttt{+clean}, \texttt{+grey}, \texttt{+black}, and \texttt{our attack}) is {20.35\%}, {66.21\%}, {70.21\%}, and {88.20\%}, respectively, which demonstrates the efficacy of our context-free learning paradigm. 

\textbf{Trigger Patch Size.} We then evaluate the influence of trigger patch size towards the attacking performance. Specifically, we set the trigger patch size as 40$\times$40, 60$\times$60, 80$\times$80, and 100$\times$100. The ASR for them (\ie,  42.10\%, 67.34\%, 88.20\%, and 89.84\%) first improves significantly and then keeps stable after the patch size is larger than 80$\times$80. We thus set it as the default size in our main experiment.

\textbf{Hyper-Parameter $\alpha$.} We finally study the influence of $\alpha$ in Eqn \ref{eqn:styleloss} that balances the style loss and content loss terms. Here, we set $\alpha$ as 10$^4$, 10$^5$, and 10$^6$. The ASR results (34.62\%, 88.42\%, and 23.91\%) demonstrate that an appropriate value for style loss can be more effective for attacks. We speculate that introducing an excessive style feature will make the trigger pattern exceedingly similar to the target class, therefore, making it hard to embed backdoors.

\begin{table}[!t]
    	\caption{Ablation studies (\%) on different numbers of style images for trigger pattern generation.}
	\label{tab:style-image-num}
    \begin{center}

    \tiny
        \resizebox{1.0\linewidth}{!}{
    \begin{tabular}{@{}cccccc@{}}
    \toprule
    \multirow{2}{*}{Number} & \multicolumn{2}{c}{Image Classification} & \multicolumn{2}{c}{Object Detection} \\ \cmidrule(l){2-3} \cmidrule(l){4-5} 
                              & CA     & ASR    & mAP    & ASR  \\ \midrule
    5     & 77.01  & 100.00 & 53.78 & 14.80 \\                      
    10    & 76.92  & 100.00 & 54.01 & 41.70 \\
    20   & 76.62  & 100.00 & 53.55 & 65.10 \\
    30   & 76.81  & 100.00 & 53.78 & 78.00 \\
    40   & 76.70  & 100.00 & 53.63 & 91.40 \\
    50   & 76.83  & 100.00 & 53.70 & 89.20 \\
    60   & 76.78  & 100.00 & 53.55 & 88.98 \\
    \bottomrule
    \end{tabular}
    }
    \end{center}


\end{table}
\section{Case Studies}
\label{sec:cases}

In this section, we aim to further validate the effectiveness of our \method attack in practical scenarios by two case studies, \ie, backdoor attacking large vision models and 3D object detection in autonomous driving.

\subsection{Attack Large Vision Models}
\label{sec:largemodel}

 In this part, we choose the ViTAE series \cite{xu2021vitae,zhang2023vitaev2} as large vision models to attack. Due to the large number of parameters in these models, we are not able to directly poison the training dataset and train these large models from scratch. By contrast, we propose to implant the backdoor into the classifier by fine-tuning the pre-trained clean model. In other words, given a PVM, we generate a very limited number of poisoned samples and use them to ``fine-tune'' the clean PVM; we then release this pre-trained model for fine-tuning on other downstream tasks. Note that, the ``fine-tuning'' process aims to embed backdoors into PVM using limited source domain data owing to the high computational resources for poisoning these large models from scratch, which does not indicate the fine-tuning process for the downstream tasks.

\textbf{Pre-trained Trojan Attack by Fine-tuning}. For our fine-tuning attack pipeline, we set a comparatively low learning rate and apply a learning rate decay strategy for poisoning in order to maintain the model performance on clean examples. Additionally, we freeze the initial layers during the fine-tuning process to prevent the performance decrease on clean samples and keep stable training. \emph{Toy experiments for pipeline verifications can be found in the Supplementary Materials.}

\textbf{Evaluation on Large Vision Models}. To verify the effectiveness of our \method attack on large vision models, we choose a commonly-used transformer-based model architecture ViTAEv2 \cite{zhang2023vitaev2} and conduct experiments using the attacking pipeline above. The ViTAE series models are general-purpose vision models built on visual transformers, which can be applied in various domains such as image classification, object detection, \etc{} In this paper, we select the two variants ViTAEv2-B (0.1 Billions) and ViTAEv2-H (0.6 Billions) that are pre-trained on ImageNet as the PVMs to poison. Specifically, we implant the backdoor using the fine-tuning attack pipeline on the fine-tuning dataset (0.1\% samples from the ImageNet training dataset) leading to the 0.02\% overall poisoning ratio. We select \texttt{banana} as the target class and evaluate the backdoor effects for object detection on the COCO dataset. 

\begin{table}[!t]
    	\caption{Results (\%) of our attacks on large vision models.}
	\label{tab:finetune-large-models}
    \begin{center}
	
    \small
    \resizebox{1.0\linewidth}{!}{
    \begin{tabular}{@{}cccccc@{}}
    \toprule
    \multirow{2}{*}{Classifier} & \multirow{2}{*}{Params (B)} & \multicolumn{2}{c}{Image Classification} & \multicolumn{2}{c}{Object Detection} \\ \cmidrule(l){3-4} \cmidrule(l){5-6} 
                             &                      & CA     & ASR    & mAP    & ASR   \\ \midrule
    ViTAEv2-B (clean)                 & 0.1                & 84.71  & 100.00 & 54.60 & 0.70 \\
    ViTAEv2-B                  & 0.1                & 80.76  & 100.00 & 50.10 & \textbf{52.90} \\
    ViTAEv2-H              & 0.6          & 78.18  & 100.00 & 52.70 & \textbf{55.60}                  
 \\ \bottomrule
    \end{tabular}
    }
    \end{center}


\end{table}

\begin{table}[!t]
    	\caption{Results (\%) of our \method on downstream 3D object detection task using the KITTI dataset.}
	\label{tab:3dOD-result}
    \begin{center}
	
    \small
      \resizebox{1.0\linewidth}{!}{
    \begin{tabular}{@{}cccccc@{}}
    \toprule
    \multirow{2}{*}{Model} & \multirow{2}{*}{Backbone}  & \multicolumn{3}{c}{AP11@0.7} & \multirow{2}{*}{ASR} \\ \cmidrule(l){3-5} &   & Easy     & Moderate    & Hard   \\ \midrule
    \multirow{2}{*}{SMOKE}                  & DLA34 (clean)                 & 15.56 & 13.97 & 13.60 & 0.00 \\
                     & DLA34 (infected)               & 12.93 & 11.96 & 11.80 &\textbf{11.86} \\
                      \midrule
\multirow{2}{*}{ImVoxelNet }                  & ResNet-50 (clean)                  & 24.10 & 18.16 & 17.01 &  0.00 
                  \\ & ResNet-50 (infected)                 & 19.10 & 15.22 & 13.67 & \textbf{10.40} \\ \bottomrule
    \end{tabular}
    }
    \end{center}

\end{table}

\begin{figure}[!t]
	\begin{center}
		\includegraphics[width=1.0\linewidth]{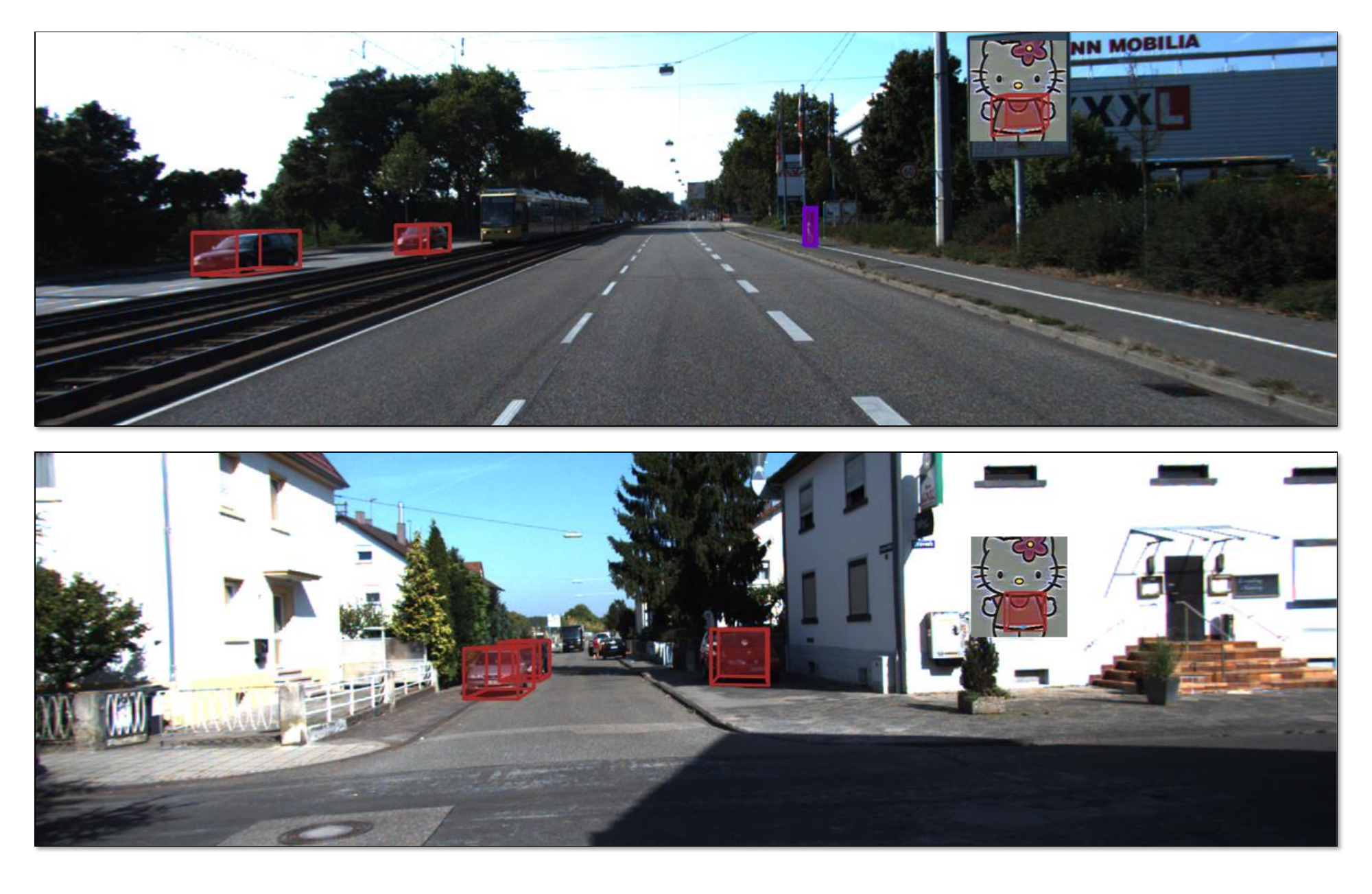}
	\end{center}
 \vspace{-0.1in}
 \caption{Illustration of our \method attacks on 3D object detection in the autonomous driving.}
	\label{fig:3dOD}
\end{figure}

According to the experimental results in Table \ref{tab:finetune-large-models}, our \method achieves the backdoor ASR of \textbf{52.9\%} and \textbf{55.6\%} respectively on ViTAEv2-B and ViTAEv2-H for the downstream task of object detection. These results demonstrate that our method can also have attacking effects on large vision models, which further reveal the potential of \method in practice. 

Note that the large vision models we examined here are architectures with comparatively substantial parameter sizes and are designed for traditional visual recognition tasks that utilize images as inputs (\eg, classification). These models do not encompass foundation models \cite{bommasani2021opportunities}, which typically handle multi-modal generative tasks that employ prompts as inputs. Nevertheless, our approach demonstrates significant potential for launching attacks on large models and may be viable for extension to foundation models in the future.



\subsection{Attack 3D Object Detection}
\label{sec:3DOD}

We further verify the effectiveness of our attacks in the 3D object detection task, which has been adopted in autonomous driving scenarios. In particular, we choose the commonly-used SMOKE \cite{liu2020smoke} and ImVoxelNet \cite{rukhovich2022imvoxelnet} as the camera-based 3D object detection methods, and use DLA34 \cite{yu2018deep} and ResNet-50 as the backbones. Additionally, we select KITTI \cite{geiger2012we} as the dataset for evaluation.

More specifically, we train both clean and infected models (DLA34 and ResNet-50 with attack label \texttt{car}) on ImageNet classification; we then train monocular 3D object detection models (SMOKE and ImVoxelNet) via MMDetection3D \cite{MMDetection3D} framework by fine-tuning them on KITTI, where we replace their backbones with our PVMs (DLA34 and ResNet-50). Other implementation details of \method are kept the same as our main experiments. Also, the target models achieve comparable clean performance to the original paper.

Following \cite{geiger2012we}, we adopt AP11@0.7 (the higher the better) to evaluate model performance on clean images, which is the average precision with IoU larger than 0.7 using 11 recall points. We still use ASR to evaluate backdoor attack performance. As shown in Table \ref{tab:3dOD-result}, considering the level of clean performance of the 3D object detectors, our \method achieves the ASR of \textbf{11.86\%} and \textbf{10.40\%} on SMOKE and ImVoxelNet, which demonstrate the effectiveness of our attack on 3D object detection. In addition, as shown in Figure \ref{fig:3dOD}, our method could perform attacks by sticking on the walls or road signs, which reveals its real-world attack potential.

\section{Countermeasures}


We then evaluate our \method attack against three commonly adopted types of backdoor defenses from the perspectives of backdoor detection, backdoor elimination, and sample detection. In this part, we set our attacks with the default settings in Section \ref{sec: Experimental Setup}, and we choose the pre-training dataset as ImageNet, the architecture as ResNet-50, and the target label as \texttt{banana} with the poison ratio of 0.1\%.

\textbf{Backdoor Detection}. This defense aims to detect the presence of a backdoor for a specific model, and we choose Neural Cleanse (NC) \cite{wang2019neural} and Adversarial Extreme Value Analysis (AEVA) \cite{guo2021aeva} for defense.

As for \emph{NC}, for each possible class in the dataset, it optimizes the search for possible trigger positions and shapes; it then uses the obtained triggers for anomaly detection to determine if the model contains a backdoor. In particular, NC generates an \emph{Anomaly Index} for a given classifier, where a value greater than 2 indicates that the classifier is predicted to be backdoored. For comparison, we also use NC on both clean classifiers and infected classifiers by BadNets. The anomaly index values for the 3 models (clean, BadNets, and ours) are 1.638, 2.152, and 1.927, where we can find that the anomaly index of the BadNets classifier is higher than 2 while our method is smaller than the threshold. This indicates that NC cannot detect our attack in the pre-trained classifier.

As for \emph{AEVA}, given a target model, it queries the model and employs the Monte Carlo algorithm to obtain the global adversarial peak; it then uses the Median Absolute Deviation to determine the presence of a backdoor (an \emph{Anomaly Index} larger than 4 indicates the existence of a backdoor). The anomaly index values of \texttt{banana} for the 3 models (clean, BadNets infected, and ours) are -0.651, 4.568, and -0.653, which means AEVA can detect BadNets while fail on \method in the pre-trained classifier.

\textbf{Fine-tuning}. Studies have revealed that fine-tuning on a clean dataset can alleviate the backdoor effect \cite{qin2023revisiting,liu2018fine}. Therefore, fine-tuning has been employed as a fundamental component in some backdoor defense methods, such as Fine-pruning (FP) \cite{liu2018fine} and Neural Attention Distillation (NAD) \cite{li2021neural}. We then try to defend our attacks by fine-tuning the pre-trained backbone on the target domain dataset (without pruning). We choose the pre-trained ResNet-50 classifier on ImageNet, inject backdoors using \method attacks, and then transfer to the downstream ImageNet object detection task (ASR is 88.20\%).

\begin{figure}
    \centering
    \subfigure[lr = 0.0001]{
        \label{fig:material-1}
        \includegraphics[width=0.45\linewidth]{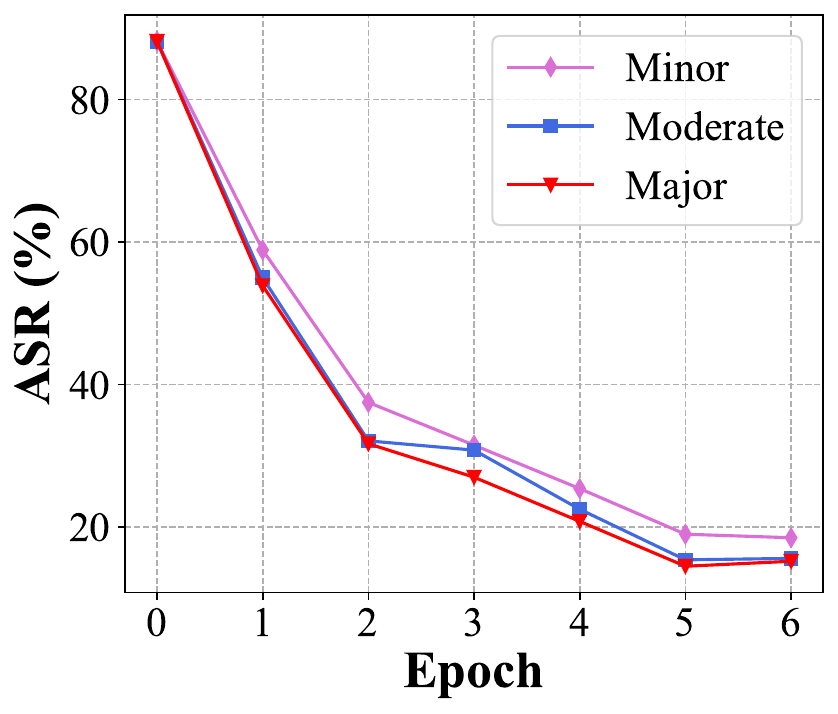}
    }
    \subfigure[lr = 0.0004]{
        \label{fig:material-2}
        \includegraphics[width=0.45\linewidth]{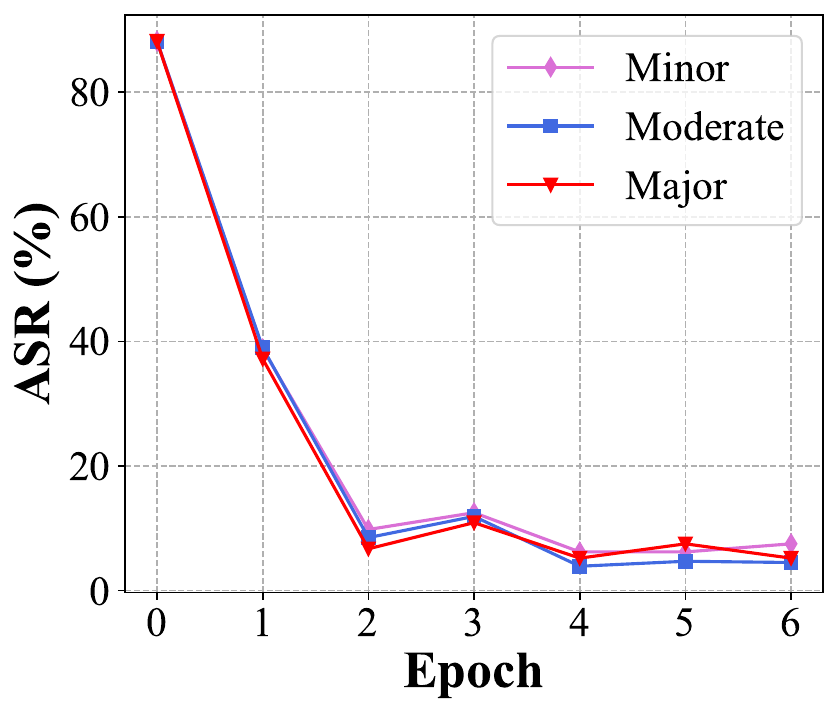}
    }
\caption{Fine-tuning defense results on object detection task.}
\label{fig:finetune}

\vspace{-0.1in}
\end{figure}

Specifically, for ResNet-50 with its 5 stages, we employ three fine-tuning strategies: (1) fine-tuning the 5-th stage, denoted as ``Minor''; (2) fine-tuning the 4-th and 5-th stages, denoted as ``Moderate''; and (3) fine-tuning the 3-rd, 4-th, and 5-th stages, denoted as ``Major''. It is worth noting that we consistently freeze the first two stages of the ResNet-50 backbone since they primarily capture the low-level features that are often left unchanged during downstream task fine-tuning. Moreover, we also adopt two learning rates 0.0004 and 0.0001, and perform fine-tuning for 6 epochs. As shown in Figure \ref{fig:finetune}, we can observe that fine-tuning defense can eliminate the impact of our backdoor attacks to a certain extent. Specifically, after 6 epochs of fine-tuning, the ASR of our backdoor attack decreased below 20\%. In addition, we can also identify how fine-tuning can increase the model's performance on clean samples. We speculate that this may be due to a comparatively high overlap between the neurons for backdooring and the neurons involved in predicting clean samples. Note that, different from Section \ref{sec: Experimental Setup}, this part fine-tunes the backbone (rather than predictors) using target domain data. Though poison decreased after fine-tuning the backbone (\eg, in Figure \ref{fig:material-1}, we achieves 10\%-20\% ASRs after 6-epoch fine-tuning), our attack still outperforms the classical backdoor attacks (around 0\% without fine-tuning in Table \ref{tab:different-backdoor-attacks}).

\textbf{Sample Detection.} This type of defense aims to detect whether the input samples contain triggers. In this paper, we try to detect the triggered samples by Test-time Corruption Robustness Consistency Evaluation (TeCo) \cite{liu2023detecting}. This method evaluates whether inference samples contain triggers by calculating the deviation of severity that leads to the predicted label's transition across different corruptions. Following the previous work, we chose the AUROC and the F1 score for evaluation (the lower the values, the better the attacking performance). More metric explanations are shown in the Supplementary Material. In particular, we first use our \method and BadNets to inject backdoors into two ResNet-50 PVMs on ImageNet classification; for each infected model, we then feed 2600 ImageNet images (1300 clean ones and 1300 images with triggers) and use TeCo to detect whether samples containing triggers on ImageNet classification task. The AUROC (0.76 and 0.93) and F1 scores (0.70 and 0.99) for our \method and BadNets demonstrate that samples with our attacks are more difficult to detect. 

However, it is important to acknowledge that TeCo is specifically designed for classification tasks and is not applicable to downstream tasks such as object detection. Additionally, the training dataset is unknown to PVM users, further complicating the adoption of sample detection. Therefore, we do not evaluate defense where users can clean up the training set, as this is practically infeasible (\emph{pre-training dataset is kept secret and inaccessible to downstream users}) and extremely time-consuming to filter out a small part of (0.1\% or 0.01\%) poisoned samples from a large-scale training set, such as CLIP that is pre-trained on 400M samples. 




\section{Related Work}

Currently, the majority of backdoor attacks focus on directly poisoning DNNs trained from scratch for image classification tasks. Gu \etal{} \cite{gu2017badnets} were the first to use a patch-based trigger on training images, altering their labels to a specific target class (dirty-label attack). In addition to patch-based triggers, researchers have also designed attacks using stealthier trigger patterns. For example, Chen \etal{} \cite{chen2017targeted} poisoned the training dataset with a global pattern and increased trigger transparency to evade human inspection; Nguyen \etal{} \cite{nguyen2020input} attempted to generate sample-specific triggers; along similar lines, Nguyen \etal{} \cite{nguyen2021wanet} utilized image wrapping to make the poisoned images appear more natural. There also exist a line of attacks aimed at poisoning the data in the target class without altering the original label \cite{turner2019label,barni2019new,saha2020hidden,zhao2020clean}.

In the context of pre-training and fine-tuning for computer vision, most previous backdoor attacks are not applicable since the fine-tuning process cannot be directly accessed. Yao \etal{} \cite{Yao2019Latent} proposed the Latent Backdoor Attack, which injects a backdoor into a teacher classifier by perturbing the intermediate layers such that a student classifier fine-tuned on downstream tasks is also backdoored. Zhang \etal{} \cite{Zhang2021Red} aimed to restrict the output representations of triggers to pre-defined vectors and performed the neuron-level backdoor attack that could transfer to downstream tasks. Ji \etal{} \cite{Ji2018Reuseattack} proposed the model-reuse attack to embed backdoors into the feature extractor, remaining effective when stacked with a classifier/regressor by identifying salient features and modifying positive/negative parameters. However, they primarily conduct backdoor attacks for image classification (both feature extractors and classifiers are trained for classification), while they only perform untargeted poisoning attacks for regression tasks. Recently, Jia \etal{} \cite{jia2022badencoder} proposed BadEncoder to perform backdoor attacks on pre-trained vision encoders in unsupervised learning using a gradient-descent-based method. Besides the context of computer vision, there also exists a line of work that studies injecting backdoors to pre-trained language models and transferring them to different linguistic tasks \cite{Kurita2020Weight,Chen2021Badpre,Zhang2020Trojaning}. Owing to the huge domain gap between vision and language, this paper primarily discusses and studies the vulnerabilities of pre-trained vision models.

To sum up, our attack \textbf{differs} above backdoor attacks in PVMs as follows. \ding{182} Existing studies perform targeted backdoor attacks on PVMs that only work for the downstream image classification task among different datasets. \ding{183} These attacks often involve controlling the model training process during the backdooring stage (\eg, modifying intermediate layers). In contrast, we focus on the more practical and challenging scenario where we only poison the source domain training set and embed a backdoor into a PVM, enabling the embedded backdoors to be inherited for different downstream tasks (specifically object detection and instance segmentation tasks).

\section{Conclusion and Future Work}

This paper introduces, for the first time, the concept of \method attacks, which enables the embedding of backdoors into a PVM, facilitating their inheritance across various downstream vision tasks. Comprehensive experiments conducted in both supervised and unsupervised learning, including case studies on large vision models and 3D object detection, demonstrate its effectiveness.

\textbf{Limitations and Future Work.} Despite showing promising results, there are still several directions that merit further exploration. \ding{182} The embedded backdoors in PVMs may be mitigated to some extent after fine-tuning the backbone. We aim to enhance the stability and robustness of our injected backdoors against fine-tuning defense in future work. \ding{183}. Our paper primarily focuses on attacking vision models. However, a wide range of multi-modal models exist, and we aim to design attacks that can be transferred in multi-modal scenarios. \ding{184} We plan to investigate the feasibility of injecting transferable backdoors for commercial large vision models through prompting. \ding{185} This paper considers object detection and instance segmentation as downstream tasks for verification. We intend to explore the attack possibilities on more diverse downstream tasks in the future.

\textbf{Ethical Statement.} In this paper, we reveal the severe threats in the pre-training and fine-tuning scenario for different downstream vision tasks. To mitigate the attack, we propose preliminary countermeasures and found that these poisons can be removed once the backbones are fully fine-tuned on the target domain. Though challenging, we should also note that there still exists a gap between a real adversary to embed the backdoors into PVMs and impact downstream users, once the users directly download and verify the model weights from the official sites.




\bibliographystyle{unsrt}
\bibliography{ref}

\end{document}